\documentclass[10pt,twocolumn,letterpaper]{article}

\usepackage{cvpr}
\usepackage{times}
\usepackage{epsfig}
\usepackage{graphicx}
\usepackage{amsmath}
\usepackage{amssymb}
\usepackage{comment}

\usepackage[pagebackref=true,breaklinks=true,letterpaper=true,colorlinks,bookmarks=false]{hyperref}

\usepackage{array}
\usepackage{boldline , multirow}
\newcommand{\PreserveBackslash}[1]{\let\temp=\\#1\let\\=\temp}
\newcolumntype{C}[1]{>{\PreserveBackslash\centering}p{#1}}
\newcolumntype{R}[1]{>{\PreserveBackslash\raggedleft}p{#1}}
\newcolumntype{L}[1]{>{\PreserveBackslash\raggedright}p{#1}}
\newcolumntype{M}[1]{>{\centering\arraybackslash}m{#1}}
\usepackage{xcolor}

\cvprfinalcopy 

\def\cvprPaperID{****} 

\ifcvprfinal\fi
\begin{document}

\title{Data augmentation to improve robustness of image captioning solutions}

\author{%
Shashank Bujimalla \\
\normalsize{Intel Corporation}\\
\texttt{\small shashankbvs@gmail.com}\\
\and
Mahesh Subedar \\
\normalsize{Intel Labs}\\
\texttt{\small mahesh.subedar@intel.com}\\
\and
Omesh Tickoo \\
\normalsize{Intel Labs}\\
\texttt{\small omesh.tickoo@intel.com}\\
}
\maketitle
\vspace{-20pt}
\begin{abstract}
In this paper, we study the impact of motion blur, a common quality flaw in real world images, on a state-of-the-art two-stage image captioning solution, and notice a degradation in solution performance as blur intensity increases. We investigate techniques to improve the robustness of the solution to motion blur using training data augmentation at each or both stages of the solution, i.e., object detection and captioning, and observe improved results. In particular, augmenting both the stages reduces the CIDEr-D degradation for high motion blur intensity from 68.7 to 11.7 on MS COCO dataset, and from 22.4 to 6.8 on Vizwiz dataset.
\end{abstract}

\section{Introduction}
Most state-of-the-art solutions for image captioning (for e.g.,~\cite{anderson2018bottom,huang2019attention}) consist of two stages. In the first stage, neural network based object detection models are trained on a large dataset and are used to generate features for salient regions in the image. In the second stage, neural network based captioning models are trained to generate captions using these region features. There exist many metrics to evaluate the accuracy of image captioning models, among which CIDEr-D is the most popular metric.

Images taken in real world are often noisy and have quality flaws. Motion blur is one such common quality flaw.
Although neural networks provide state-of-the-art results, they have been shown to have issues with noisy or out-of-distribution data. In this paper, we study the impact of motion blur on a state-of-the-art image captioning solution and notice a degradation in CIDEr-D scores as motion blur intensity increases. We investigate techniques to improve their robustness using data augmentation. Although training data augmentation is a popular technique to improve model performance, it has not been usually studied in image captioning solutions, most likely due to their two-stage solution nature; also, there are no studies on using it to improve their robustness to motion blur. We use MS COCO~\cite{chen2015microsoft} and Vizwiz~\cite{gurari2020captioning} image captioning datasets in this study.

\begin{table*}[!tb]
\small
\renewcommand{\arraystretch}{1.3}
\begin{center}
\begin{tabular}{M{3 cm}|C{0.7 cm} C{0.7 cm} C{0.7 cm} C{0.7 cm} C{0.001cm} | C{0.7 cm} C{0.7 cm} C{0.7 cm} C{0.7 cm} | C{1.4 cm} C{1 cm}}
& \multicolumn{4}{c}{MS COCO Dataset} && \multicolumn{6}{c}{Vizwiz Dataset}   \\ \cline{2-12}
Training approach & MB0 & MB1 & MB2 & MB3 && MB0 & MB1 & MB2 & MB3 & With blur & No blur \\ \hline
No-Aug & 117.1 & 111.4 & 95.0 & 48.4 && 48.8 & 47.0 & 40.9 & 26.4 & 47.2 & 53.0 \\
ObjDet-Aug & 116.6 & 114.6 & 111.7 & 100.2 && 48.9 & 48.1 & 45.6 & 39.5 & 47.0 & 53.3 \\
Cap-Aug & 116.8 & 115.0 & 108.8 & 85.1 && 50.0 & 49.2 & 46.9 & 38.2 & \textbf{49.0} & 53.2 \\
ObjDet-Cap-Aug & \textbf{117.4} & \textbf{116.0} & \textbf{113.4} & \textbf{105.7} && \textbf{50.3} & \textbf{49.9} & \textbf{48.1} & \textbf{43.5} & 48.9 & \textbf{54.1} \\
\end{tabular}
\end{center}
\caption{\small Impact of additional motion blur (denoted as MB0, MB1, MB2, and MB3 for intensities 0, 1, 2, and 3 respectively) on CIDEr-D scores of MS COCO and VizWiz dataset validation splits. Our techniques ("ObjDet-Aug", "Cap-Aug", "ObjDet-Cap-Aug") consistently improve the CIDEr-D robustness to motion blur (MB1, MB2, MB3) compared to the "No-Aug" technique. Also shown is impact of our techniques on Vizwiz validation split images with and without blur; they show improved CIDEr-D compared to the "No-Aug" technique.}
\vspace*{-8pt}
\label{tab:all_results}
\end{table*}

\section{Our approach}
\label{sec:our_app}
As proposed in the popular two-stage approach~\cite{anderson2018bottom}, we first train a Faster R-CNN object detection model (we use ResNeXt-152 backbone) on Visual genome dataset, and extract region proposal features for the images in target image captioning dataset, i.e., MS COCO or Vizwiz. We then train the AoANet image captioning model~\cite{huang2019attention} using these features and cross-entropy loss for 25 epochs with a minibatch size of 50. (We do not perform Self Critical Sequence Training step to further improve the CIDEr-D score, but we expect that the trends in our results would not change with that additional step).
To model additional global motion blur on the images (examples are shown in Appendix), we use OpenCV image blur function with tap sizes (6,1), (18,6), (45,12) to represent blur intensity levels low, medium and high (referred as 1, 2 and 3 hereafter in this paper) respectively. We refer to the original images in the datasets as having 0 additional motion blur.

During inference phase, we extract region proposal features for images with additional motion blur intensities of 0, 1, 2 and 3 using the trained object detection model; we then pass them to the trained captioning model to infer captions for their respective motion blur intensities.
During training phase, we perform data augmentation wherein images with additional motion blur are used during model training with a chosen probability. We study four techniques based on the stage(s) of the solution at which data augmentation is done; we refer to them as “No-Aug”, “ObjDet-Aug”, “Cap-Aug”, and “ObjDet-Cap-Aug”. In "No-Aug", which is the default technique, the images with additional motion blur are not used during training phase. In “ObjDet-Aug”, we augment the images with additional motion blur only during object detection model training. In “Cap-Aug”, we do not augment object detection model training, but extract the region proposal features for additional motion blur intensities and augment only the captioning model training. In “ObjDet-Cap-Aug”, we perform data augmentations during training of both object detection and captioning models.

To augment training of object detection model, we use images with additional motion blur intensities 0, 1 and 2 with 0.8, 0.1 and 0.1 probabilities respectively (blur intensity 3 is not used). To augment training of captioning model, we extract region proposal features for images with additional motion blur intensities 0, 1, 2 and 3, and then use them during captioning model training with 0.5, 0.2, 0.2 and 0.1 probabilities respectively. 
Note: Our choice of these probabilities is empirically based on a few trials (i.e., not a rigorous search) so that there is no significant degradation in model accuracy (i.e., mAP and CIDEr-D) for images with 0 additional motion blur; they can also be chosen based on the expected distribution of noise in target dataset.

\section{Results}
In Table~\ref{tab:all_results}, we show the results (CIDEr-D) using our training techniques on both MS COCO and Vizwiz datasets.
We evaluate our solution on validation splits for MS COCO (Karpathy split, i.e., 5000 images) and Vizwiz (~\cite{gurari2020captioning} split, i.e., 7542 images) datasets.
Qualitative examples of our results are shown in Appendix (Figure~\ref{fig:qual-ex1}). We denote additional motion blur intensity levels 0, 1, 2 and 3 as MB0, MB1, MB2 and MB3 respectively. As motion blur intensity increases (MB0 through MB3), solutions trained with our augmentation techniques ("ObjDet-Aug", "Cap-Aug", "ObjDet-Cap-Aug") have higher CIDEr-D than the ones with no augmentation (No-Aug). Also, solutions with augmentation at both the training stages ("ObjDet-Cap-Aug") are more robust (higher MB1 through MB3 scores) than the ones with augmentation at only one of the training stages ("ObjDet-Aug", "Cap-Aug"). The slight degradation in MB0 scores for "ObjDet-Aug", "Cap-Aug" on MS COCO could most likely be fixed with a better choice of augmentation probabilities mentioned in Section~\ref{sec:our_app}.
In Table~\ref{tab:all_results}, we also separate the original images in Vizwiz dataset (i.e., MB0), which were taken by visually challenged, into images with blur ({\raise.17ex\hbox{$\scriptstyle\sim$}}4.5K images) and without blur ({\raise.17ex\hbox{$\scriptstyle\sim$}}3K images), wherein the blur annotations were done by crowd workers. We observe improved CIDEr-D for both types of images using our techniques.

We analyzed the effect of motion blur on the number of region proposal features obtained from object detection model. Histograms at different blur intensities (MB0 through MB3) are presented in Appendix Figure~\ref{fig:MSCOCO-fea-count} \& \ref{fig:vizwiz-fea-count} for MS COCO and Vizwiz respectively. When object detection model training is not augmented with additional motion blur images, the number of region features decrease (Figure~\ref{fig:MSCOCO-fea-count}a, \ref{fig:vizwiz-fea-count}a) as the blur intensity increases (MB0 through MB3); this correlates with the decrease in CIDEr-D for "No-Aug" and "Cap-Aug" techniques (MB0 through MB3) in Table~\ref{tab:all_results}; since the "Cap-Aug" technique uses the MB1 through MB3 region features during captioning model training, its decrease in CIDEr-D is moderate compared to "No-Aug" technique. After augmenting the images with additional motion blur during object detection model training, the number of region features do not decrease significantly (Figure~\ref{fig:MSCOCO-fea-count}b, \ref{fig:vizwiz-fea-count}b) as the blur intensity increases (MB0 through MB3); this correlates with the relatively less decrease in CIDEr-D for "ObjDet-Aug" and "ObjDet-Cap-Aug" techniques (MB0 through MB3) in Table~\ref{tab:all_results}.

Our studies should motivate research towards augmenting image captioning models with quality flaws seen in real world images. Also, they show that augmenting both the object detection and captioning models is critical to improve the robustness of the overall solution.

{
\bibliographystyle{ieee_fullname}
\bibliography{dataaug_cvpr21vw}
}
\begin{figure*}
\centering
\includegraphics[width=1.0\linewidth]{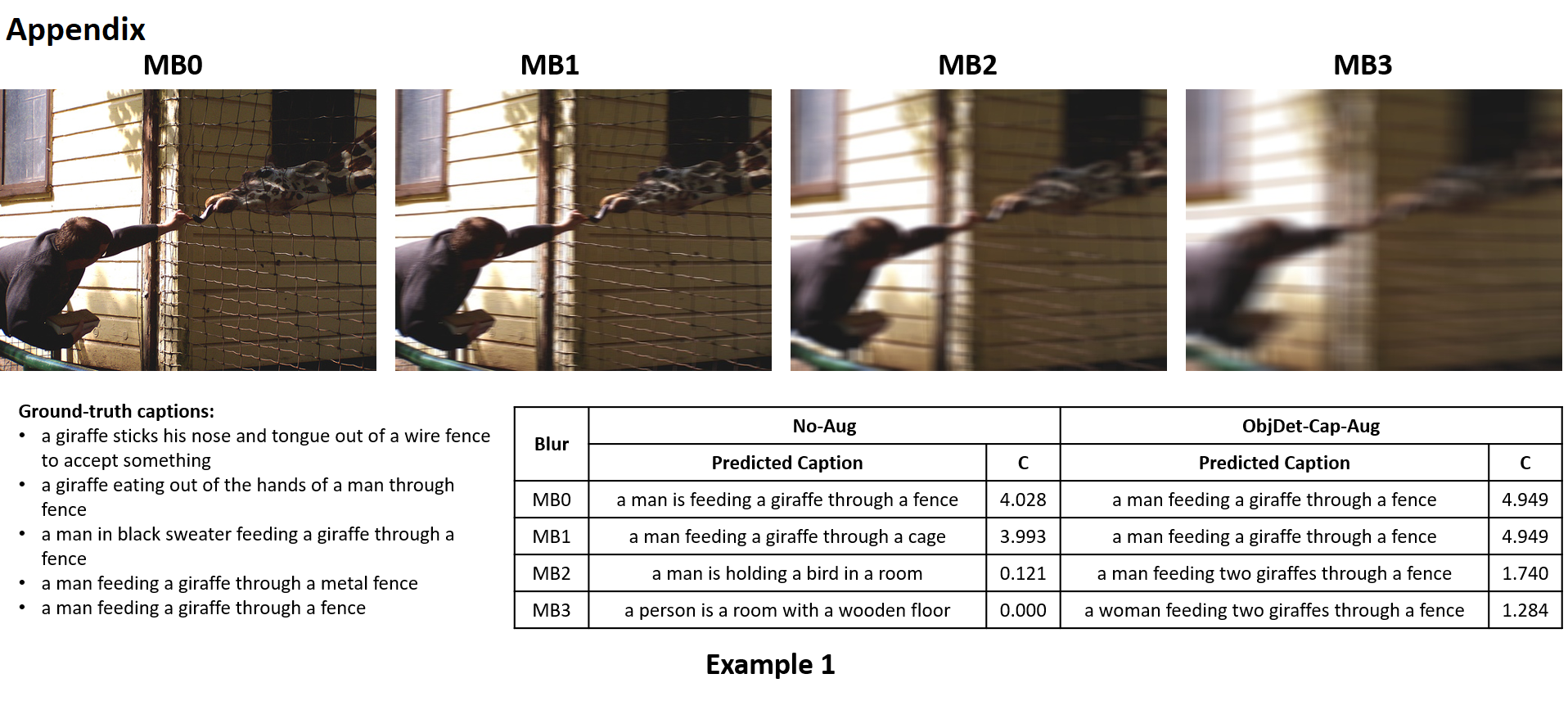}
\includegraphics[width=1.0\linewidth]{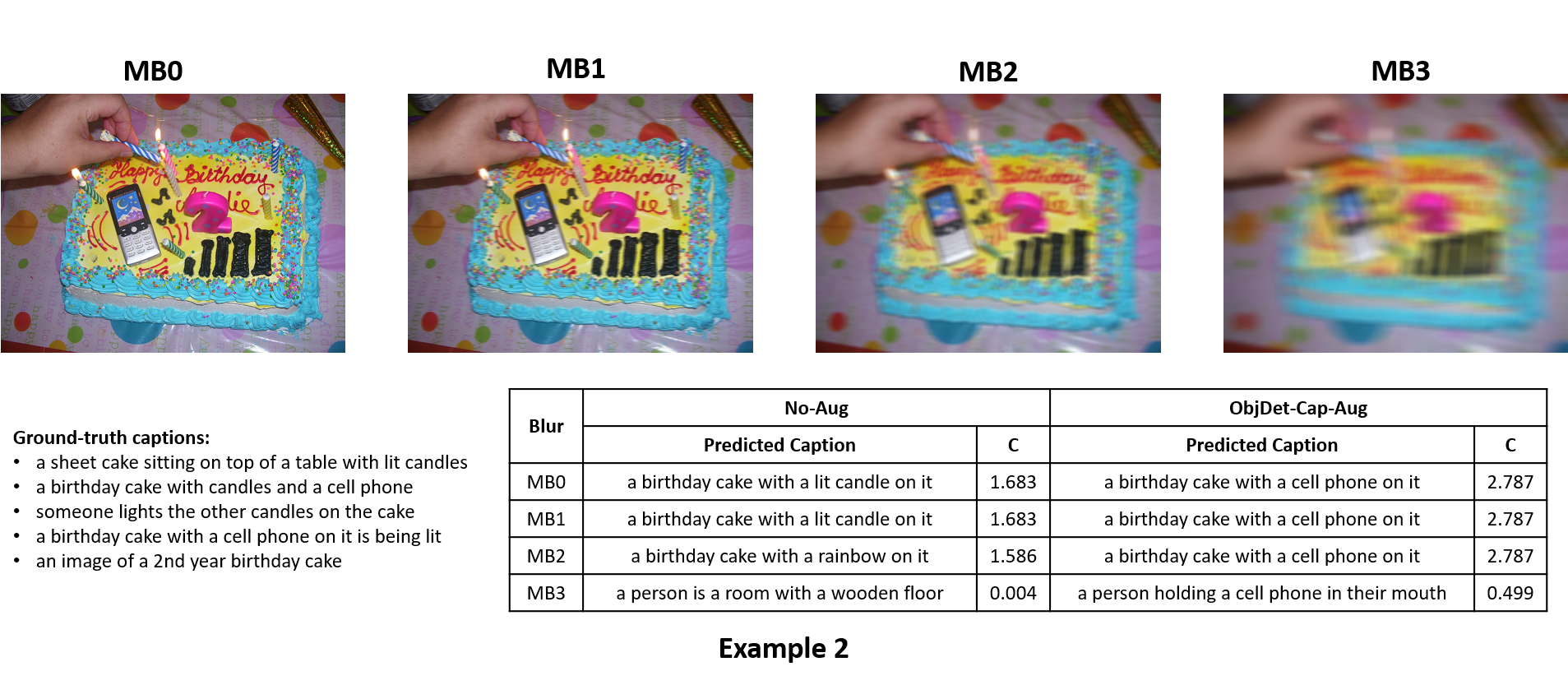}
\includegraphics[width=1.0\linewidth]{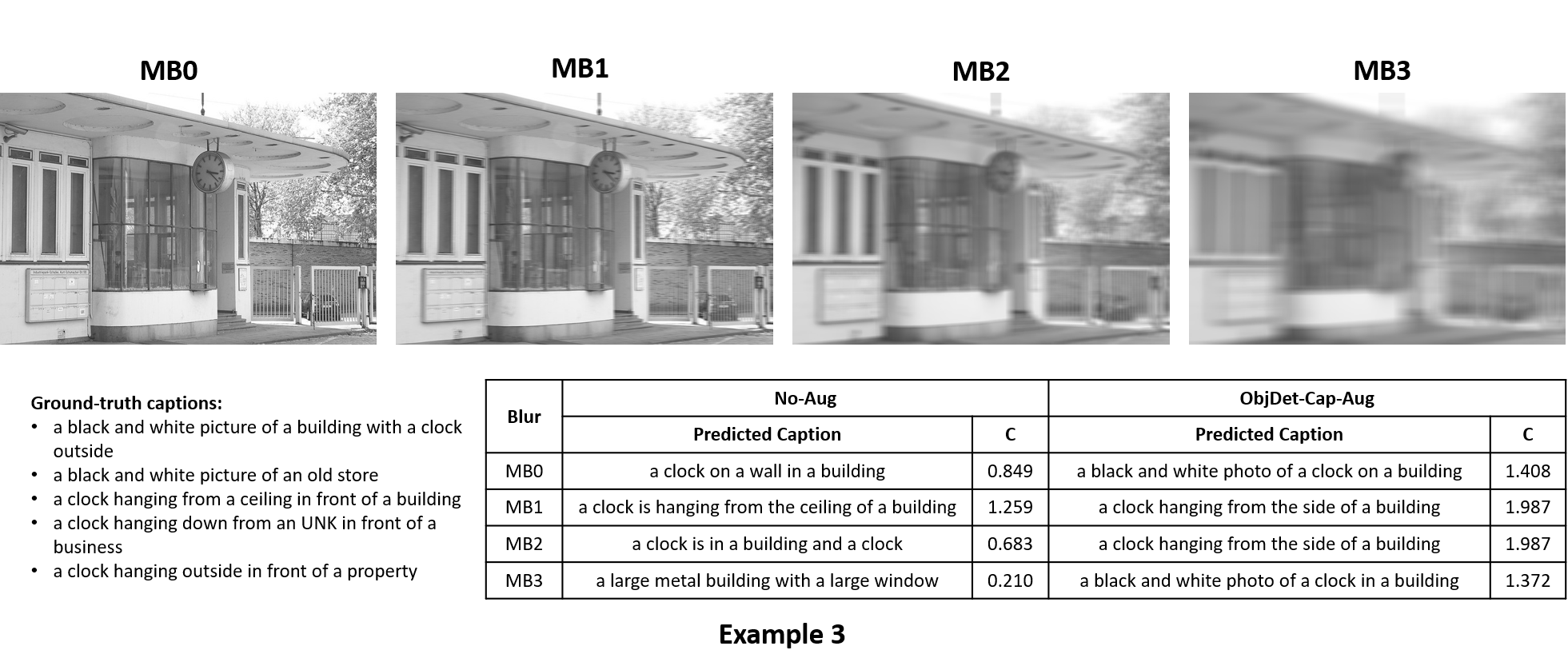}
\caption{\small
Example images (MB0) and their ground-truth captions from MS COCO Karpathy val split. Also shown are their images with additional motion blur (MB1, MB2, MB3) along with their respective predicted captions and CIDEr-D scores (C) using No-Aug, ObjDet-Cap-Aug techniques.
}
\label{fig:qual-ex1}
\end{figure*}

\begin{figure*}
\centering
\includegraphics[width=1.0\linewidth]{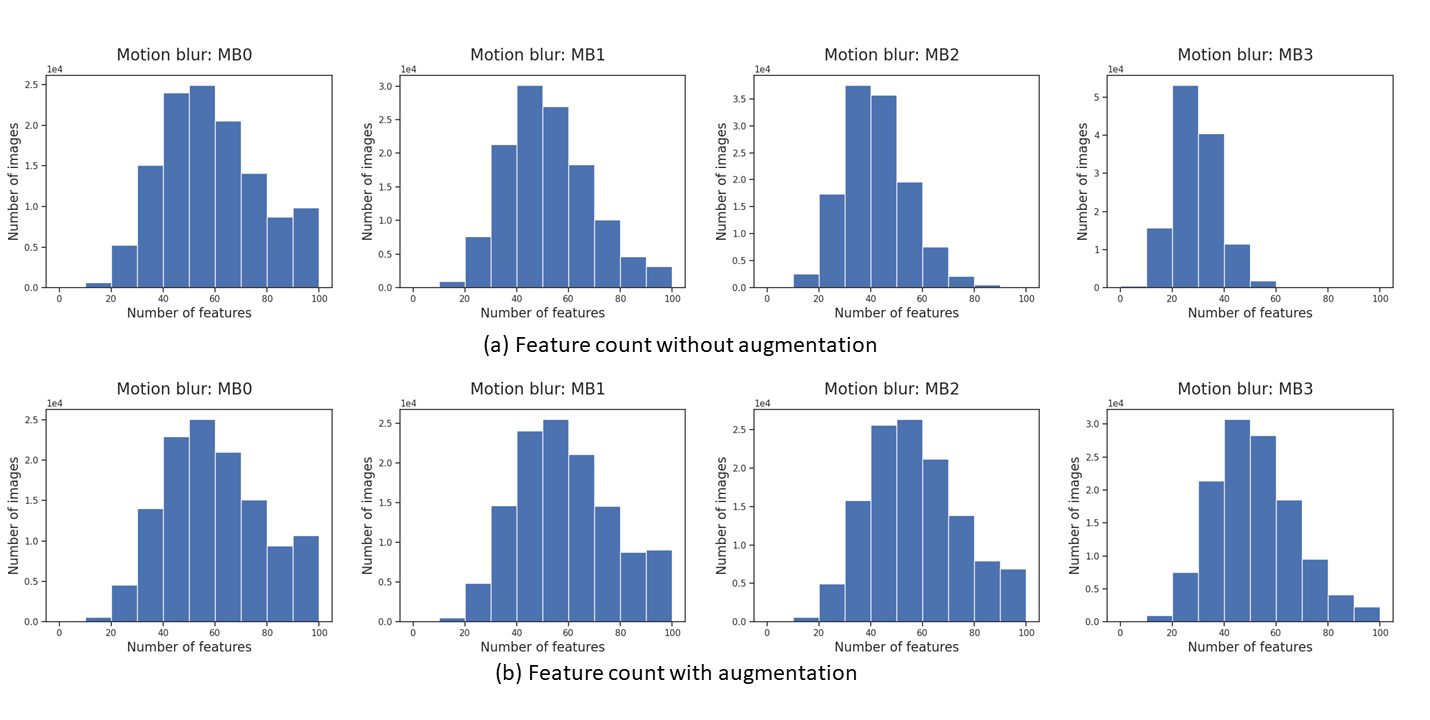}
\caption{\small
Histograms of the number of region proposal features obtained from object detection model for MS COCO dataset. The horizontal bins are the number of region proposal features and the height of the histogram shows the image count. In (a), the object detection model is not augmented with motion-blur images; the number of region proposal features decrease as the motion blur intensity increases (MB0 through MB3). In (b), the object detection model is augmented with motion-blur images; the number of region proposal features do not significantly decrease as the motion blur intensity increases (MB0 through MB3).
}
\label{fig:MSCOCO-fea-count}
\end{figure*}
\begin{figure*}
\centering
\includegraphics[width=1.0\linewidth]{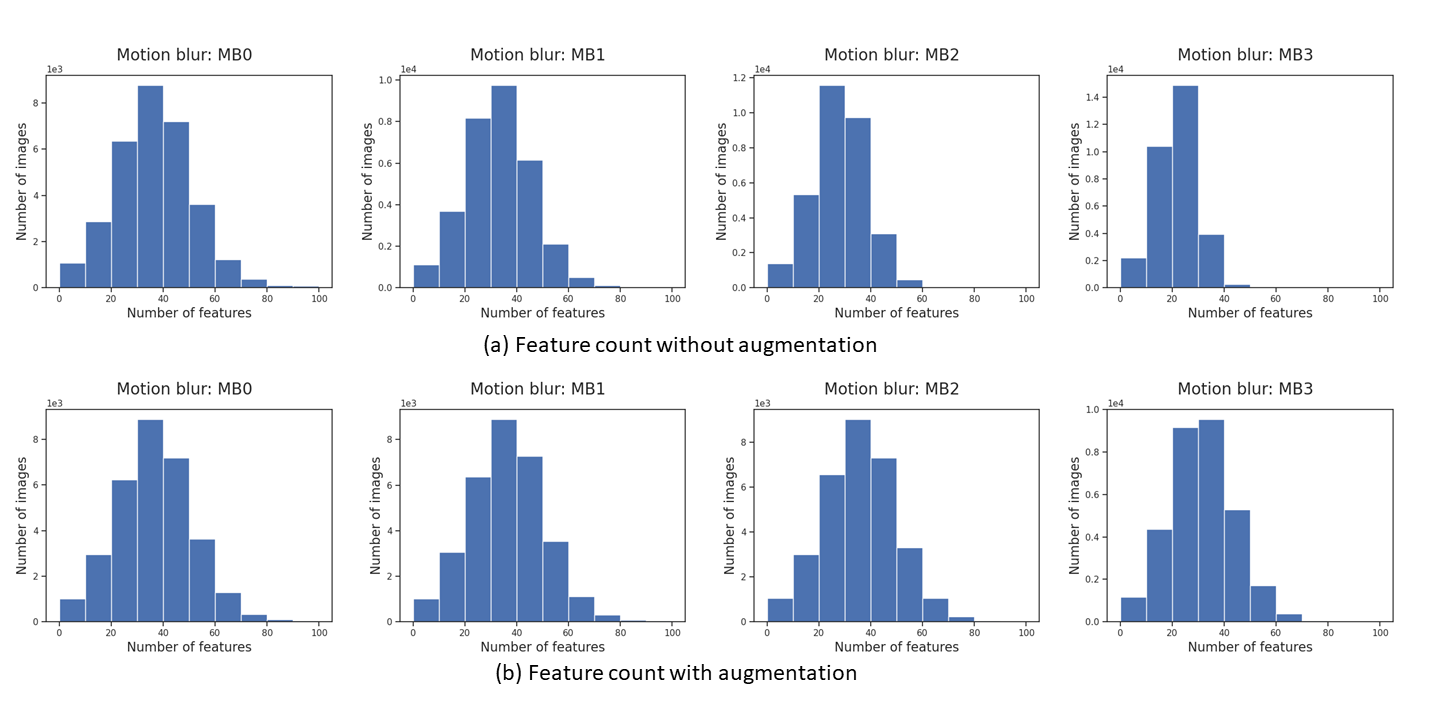}
\caption{\small
Histograms of the number of region proposal features obtained from object detection model for Vizwiz dataset. The horizontal bins are the number of region proposal features and the height of the histogram shows the image count. In (a), the object detection model is not augmented with motion-blur images; the number of region proposal features decrease as the motion blur intensity increases (MB0 through MB3). In (b), the object detection model is augmented with motion-blur images; the number of region proposal features do not significantly decrease as the motion blur intensity increases (MB0 through MB3).
}
\label{fig:vizwiz-fea-count}
\end{figure*}
\end{document}